\documentclass[11pt,twocolumn]{article}
\usepackage{authblk}
\usepackage[hyperref]{emnlp2018}

\usepackage{times}
\usepackage{url}
\usepackage{paralist}
\usepackage{latexsym}
\usepackage{ucs}
\usepackage{fontenc}
\usepackage[utf8x]{inputenc}
\usepackage[english]{babel}
\usepackage{dsfont}
\usepackage{textcomp}
\usepackage{amsfonts}
\usepackage{amsmath}
\usepackage{amssymb}
\usepackage{graphicx}
\usepackage{adjustbox}
\usepackage{sidecap}
\usepackage{pgfplots}
\pgfplotsset{compat=1.10}
\usepackage{array} % used in projection illustration and table row numbering
\usepackage{algorithm}
\usepackage[noend]{algpseudocode}
\usepackage{multirow}
\usepackage{bm}
\usepackage{breqn}
\usepackage{booktabs}
\usepackage{upgreek}
\usepackage{stackrel}
\usepackage{comment}
\usepackage{cleveref}

\usepackage{tikz}
\usetikzlibrary{arrows}
\newcommand*\circled[1]{\tikz[baseline=(char.base)]{
            \node[shape=circle,draw,inner sep=1pt] (char) {$#1$};}}

\aclfinalcopy

\makeatletter
\newcommand{\@emptybiblabel}[1]{}
\newcommand\xleftrightarrow[2][]{%
  \ext@arrow 9999{\longleftrightarrowfill@}{#1}{#2}}
\newcommand\longleftrightarrowfill@{%
  \arrowfill@\leftarrow\relbar\rightarrow}
\makeatother

\newcommand{\com}[1]{}

\newcommand{\mg}[0]{multigraph}
\newcommand{\WN}[0]{WordNet}
\newcommand{\ergm}[0]{ERGM}

\newcommand{\todo}[1]{}
\newcommand{\jacob}[1]{}
\newcommand{\yuval}[1]{}

\renewcommand{\vec}[1]{\bm{#1}}

\newcommand{\ve}{\ensuremath \vec{e}}

\newcommand{\tneg}[0]{\ensuremath \tilde{t}}
\newcommand{\Gneg}[0]{\ensuremath \tilde{G}}
\newcommand{\mscore}[0]{\ensuremath \psi_{\footnotesize{\textsc{ergm}+}}}
\newcommand{\escore}[0]{\ensuremath \psi_{\footnotesize{\textsc{ergm}}}}

\newcommand{\sysname}[0]{\textsc{m3gm}}

\title{Predicting Semantic Relations using Global Graph Properties}

\author{\textbf{Yuval Pinter} and \textbf{Jacob Eisenstein}\\
		School of Interactive Computing\\
		Georgia Institute of Technology, Atlanta, GA\\
        \{uvp, jacobe\}@gatech.edu}

\date{}

\begin{document}

\maketitle
\begin{abstract}
Semantic graphs, such as \WN{}, are resources which curate natural language on two distinguishable layers.
On the local level, individual relations between synsets (semantic building blocks) such as hypernymy and meronymy enhance our understanding of the words used to express their meanings.
Globally, analysis of graph-theoretic properties of the entire net sheds light on the structure of human language as a whole.
In this paper, we combine global and local properties of semantic graphs through the framework of Max-Margin Markov Graph Models (\sysname{}), a novel extension of Exponential Random Graph Model (\ergm{}) that scales to large multi-relational graphs.
We demonstrate how such global modeling improves performance on the local task of predicting semantic relations between synsets, yielding new state-of-the-art results on the WN18RR dataset, a challenging version of WordNet link prediction in which ``easy'' reciprocal cases are removed.
In addition, the \sysname{} model identifies multirelational motifs that are characteristic of well-formed lexical semantic ontologies.

\end{abstract}

\section{Introduction}

Semantic graphs, such as \WN{} \cite{wordnet}, encode the structural qualities of language as a representation of human knowledge.
On the local level, they describe connections between specific semantic concepts, or \textbf{synsets}, through individual edges representing relations such as hypernymy (`is-a') or meronymy (`is-part-of');
on the global level, they encode emergent regular properties in the induced relation graphs.
Local properties have been subject to extensive study in recent years via the task of \textbf{relation prediction}, where individual edges are found based mostly on distributional methods that embed synsets and relations into a vector space~\cite[e.g.][]{SocherChenManningNg2013,bordes2013translating,toutanovachen2015,neelakantan2015compositional}.
In contrast, while the structural regularity and significance of global aspects of semantic graphs is well-attested~\cite{sigman2002global}, global properties have rarely been used in prediction settings.
In this paper, we show how global semantic graph features can facilitate in local tasks such as relation prediction.

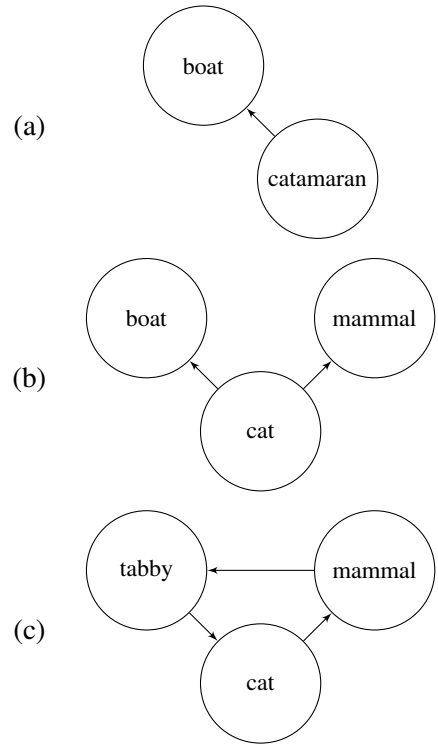
\begin{figure}
  \centering
  \begin{tabular}{cc}
    & \multirow{7}{*}{
      \begin{tikzpicture}
        \small
        \tikzset{vertex/.style = {shape=circle,draw,minimum size=5em}}
        \tikzset{edge/.style = {->,> = latex'}}
        \node[vertex] (a) at  (0,0) {catamaran};
        \node[vertex] (c) at  (-1.5,1.5) {boat};
        %edges
        \draw[edge] (a) to (c); % suggested
      \end{tikzpicture}} \\
      & \\
      & \\
      (a) & \\
      & \\
      & \\
      & \\
    & \multirow{7}{*}{
      \begin{tikzpicture}
        \small
        \tikzset{vertex/.style = {shape=circle,draw,minimum size=5em}}
        \tikzset{edge/.style = {->,> = latex'}}
        \node[vertex] (a) at  (0,0) {cat};
        \node[vertex] (b) at  (1.5,1.5) {mammal};
        \node[vertex] (c) at  (-1.5,1.5) {boat};
        %edges
        \draw[edge] (a) to (b);
        \draw[edge] (a) to (c); % suggested
      \end{tikzpicture}} \\
      & \\
      & \\
      (b) & \\
      & \\
      & \\
      & \\
    & \multirow{7}{*}{
      \begin{tikzpicture}
        \small
        \tikzset{vertex/.style = {shape=circle,draw,minimum size=5em}}
        \tikzset{edge/.style = {->,> = latex'}}
        \node[vertex] (a) at  (0,0) {cat};
        \node[vertex] (b) at  (1.5,1.5) {mammal};
        \node[vertex] (c) at  (-1.5,1.5) {tabby};
        %edges
        \draw[edge] (a) to (b);
        \draw[edge] (c) to (a);
        \draw[edge] (b) to (c); % suggested
      \end{tikzpicture}} \\
      & \\
      & \\
      (c) & \\
      & \\
      & \\
      & \\
  \end{tabular}
  \caption{\label{fig:motifs} Probable (a) and improbable (b-c) structures in a hypothetical hypernym graph.}
\end{figure}

To motivate this approach, consider the hypothetical hypernym graph fragments in \Cref{fig:motifs}: in (a), the semantic concept (synset) `catamaran' has a single hypernym, `boat'. This is a typical property across a standard hypernym graph.
	In (b), the synset `cat' has two hypernyms, an unlikely event.
While a local relation prediction model might mistake the relation between `cat' and `boat' to be plausible, for whatever reason, a high-order graph-structure-aware model should be able to discard it based on the knowledge that a synset should not have more than one hypernym.
In (c), an impossible situation arises: a cycle in the hypernym graph leads each of the participating synsets to be predicted by transitivity as its own hypernym, contrary to the relation's definition.
However, a purely local model has no explicit mechanism for rejecting such an outcome.

In this paper, we examine the effect of global graph properties on the link structure via the \WN{} relation prediction task.
Our hypothesis is that features extracted from the entire graph can help constrain local predictions to structurally sound ones~\cite{guo2007recovering}.
Such features are often manifested as aggregate counts of small subgraph structures, known as \textbf{motifs}, such as the number of nodes with two or more outgoing edges, or the number of cycles of length 3.
Returning to the example in \Cref{fig:motifs}, each of these features will be affected when graphs (b) and (c) are evaluated, respectively.

To estimate weights on local and global graph features, we build on the \textbf{Exponential Random Graph Model} (\ergm), a log-linear model over networks utilizing global graph features~\cite{holland1981exponential}.
In \ergm{}s, the likelihood of a graph is computed by exponentiating a weighted sum of the features, and then normalizing over all possible graphs.
This normalization term grows exponentially in the number of nodes, and in general cannot be decomposed into smaller parts.
Approximations are therefore necessary to fit \ergm{}s on graphs with even a few dozen nodes, and the largest known \ergm{}s scale only to thousands of nodes~\citep{schmid2017exponential}.
This is insufficient for \WN{}, which has an order of $10^5$ nodes.

We extend the \ergm{} framework in several ways.
First, we replace the maximum likelihood objective with a margin-based objective, which compares the observed network against alternative networks;
we call the resulting model the Max-Margin Markov Graph Model (\sysname{}), drawing on ideas from structured prediction~\citep{MMMN}.
The gradient of this loss is approximated by importance sampling over candidate negative edges, using a local relational model as a proposal distribution.
The complexity of each epoch of estimation is thus linear in the number of edges, making it possible to scale up to  the $10^5$ nodes in \WN{}.\footnote{Although in principle the number of edges could grow quadratically with the number of nodes, \newcite{steyvers2005large} show that semantic graphs like WordNet tend to be very sparse, so that the number of observed edges grows roughly linearly with the number of nodes.}
Second, we address the multi-relational nature of semantic graphs, by incorporating a combinatorial set of labeled motifs.
Finally, we link graph-level relational features with distributional information, by combining the \sysname{} with a dyad-level model over word sense embeddings.

We train \sysname{} as a re-ranker, which we apply to a a strong local-feature baseline on the WN18RR dataset~\cite{dettmers2018conve}. This yields absolute improvements of 3-4 points on all commonly-used metrics. 
Model inspection reveals that \sysname{} assigns importance to features from all relations, and captures some interesting inter-relational properties that lend insight into the overall structure of \WN{}.\footnote{Our code is available at \url{http://www.github.com/yuvalpinter/m3gm}.}

\section{Related Work}

\paragraph{Relational prediction in semantic graphs.} Recent approaches to relation prediction in semantic graphs generally start by embedding the semantic concepts into a shared space and modeling relations by some operator that induces a score for an embedding pair input.
We use several of these techniques as base models~\cite{nickel2011three,bordes2013translating,yang2014embedding}; detailed description of these methods is postponed to \Cref{ssec:local}.
\newcite{SocherChenManningNg2013} generalize over the approach of \newcite{nickel2011three}
by using a bilinear tensor which assigns multiple parameters for each relation; \newcite{shi2017proje} project the node embeddings in a translational model similar to that of \newcite{bordes2013translating}; \newcite{dettmers2018conve} apply a convolutional neural network by reshaping synset embeddings to 2-dimensional matrices. None of these embedding-based approaches incorporate structural information; in general, improvements in embedding-based methods are expected to be complementary to our approach.

Some recent works compose single edges into more intricate motifs, such as \newcite{guu-miller-liang:2015:EMNLP},
who define a task of \textbf{path prediction}
and compose various functions to solve it.
They find that compositionalized bilinear models perform best on \WN{}.
\newcite{minervini2017adversarial} train link-prediction models against an adversary that produces examples which violate structural constraints such as symmetry and transitivity.
Another line of work builds on local neighborhoods of relation interactions and automatic detection of relations from syntactically parsed text~\cite{riedel2013relation,toutanova2015representing}.
\newcite{gcn} use Graph Convolutional Networks to predict relations while considering high-order neighborhood properties of the nodes in question.
In general, these methods aggregate information over local neighborhoods, but do not explicitly model structural motifs.

Our model introduces interaction features between relations (e.g., hypernyms and meronyms) for the goal of relation prediction.
To our knowledge, this is the first time that relation interaction is explicitly modeled into a relation prediction task.
Within the \ergm{} framework, \newcite{lu2010supervised} train a limited set of combinatory path features for social network link prediction.

\paragraph{Scaling exponential random graph models.} The problem of approximating the denominator of the \ergm{} probability has been an active research topic for several decades.
Two common approximation methods exist in the literature.
In \textbf{Maximum Pseudolikelihood Estimation}~\cite[\textbf{MPLE};][]{strauss1990pseudolikelihood}, a graph's probability is decomposed into a product of the probability for each edge, which in turn is computed based on the \ergm{} feature difference between the graph excluding the edge and the full graph.
\textbf{Monte Carlo Maximum Likelihood Estimation}~\cite[\textbf{MCMLE};][]{snijders2002markov} follows a sampling logic, where a large number of graphs is randomly generated from the overall space under the intuition that the sum of their scores would give a good approximation for the total score mass.
The probability for the observed graph is then estimated following normalization conditioned on the sampling distribution, and its precision increases as more samples are gathered.
Recent work found that applying a parametric bootstrap can increase the reliability of MPLE, while retaining its superiority in training speed~\cite{schmid2017exponential}.
Despite this result, we opted for an MCMLE-based approach for \sysname{}, mainly due to the ability to keep the number of edges constant in each sampled graph.
This property is important in our setup, since local edge scores added or removed to the overall graph score can occasionally dominate the objective function, giving unintended importance to the overall edge count.

\section{Max-Margin Markov Graph Models}
\label{sec:ergm}
Consider a graph $G = (V,E)$, where $V$ is a set of vertices and $E = \{(s_i,t_i)\}_{i=1}^{|E|}$ is a set of directed edges. The \ergm{} scoring function defines a probability over $\mathcal{G}_{|V|}$, the set of all graphs with $|V|$ nodes. This probability is defined as a log-linear function,
\begin{equation}
\label{eq:score}
P_{\text{ERGM}}(G) \propto \escore(G) = \exp{}\left(\boldsymbol{\theta}^T \mathbf{f}(G)\right),
\end{equation}
where $\mathbf{f}$ is a feature function, from graphs to a vector of feature counts. Features are typically counts of motifs --- small subgraph structures --- as described in the introduction. The vector $\boldsymbol{\theta}$ is the parameter to estimate.

In this section we discuss our adaptation of this model to the domain of semantic graphs,
leveraging their idiosyncratic properties. Semantic graphs are composed of multiple relation types, which the feature space needs to accommodate; their nodes are linguistic constructs (semantic concepts) associated with complex interpretations, which can benefit the graph representation through incorporating their embeddings in $\mathds{R}^d$ into a new scoring model.
We then present our \sysname{} framework to perform reliable and efficient parameter estimation on the new model.

\subsection{Graph Motifs as Features}
\label{sec:feat-eng}

Based on common practice in \ergm{} feature extraction~\cite[e.g.,][]{morris2008specification},
we select the following graph features as a basis:
\begin{itemize}
  \setlength\itemsep{0pt}
\item Total edge count;
\item Number of cycles of length $k$, for $k \in \{2,3\}$;
\item Number of nodes with exactly $k$ outgoing (incoming) edges, for $k \in \{1,2,3\}$;
\item Number of nodes with \textbf{at least} $k$ outgoing (incoming) edges, for $k \in \{1,2,3\}$;
\item Number of paths of length 2;
\item Transitivity: the proportion of length-2 paths $u\rightarrow v \rightarrow w$ where an edge $u \rightarrow w$ also exists.
\end{itemize}

Semantic graphs are \mg s, where multiple relationships (hypernymy, meronymy, derivation, etc.) are overlaid atop a common set of nodes.
For each relation $r$ in the relation inventory $\mathcal{R}$, we denote its edge set as $E_r$, and redefine $E = \bigcup_{r\in\mathcal{R}} E_r$, the union of all labeled edges.
Some relations do not produce a connected graph, while others may coincide with each other frequently, possibly in regular but intricate patterns: for example, derivation relations tend to occur between synsets in the higher, more abstract levels of the hypernym graph. 
We represent this complexity by expanding the feature space to include relation-sensitive \textbf{combinatory motifs}.
For each feature template from the basis list above, we extract features for all possible combinations of relation types existing in the graph.
Depending on the feature type, these could be relation singletons, pairs, or triples; they may be order-sensitive or order-insensitive.
For example:
\begin{itemize}
  \item A combinatory `transitivity' feature will be extracted for the proportion of paths $u \xrightarrow[]{hypernym} v \xrightarrow[]{meronym} w$ where an edge $u \xrightarrow[]{has\_part} w$ also exists.
  \item A combinatory `2-outgoing' feature will be extracted for the number of nodes with exactly one \textit{derivation} and one \textit{has\_part}.
  \end{itemize}
The number of features thus scales in $O(|\mathcal{R}|^K)$
for a feature basis which involves up to $K$ edges in any feature,
and so our 17 basis features (with $K=3$) generate a combinatory feature set with roughly 3,000 features for the 11-relation version of \WN{} used in our experiments (see \Cref{ssec:wnrr}).

\subsection{Local Score Component}
\label{ssec:local}
In classical \ergm{} application domains such as social media or biological networks, nodes tend to have little intrinsic distinction, or at least little meaningful intrinsic information that may be extracted prior to applying the model.
In semantic graphs, however, the nodes represent synsets, which are associated with information that is both valuable to predicting the graph structure and approximable using unsupervised techniques such as embedding into a common $d$-dimensional vector space based on copious amounts of available data.
We thus modify the traditional scoring function from \cref{eq:score} to include node-specific information, by introducing a relation-specific
\textbf{association operator} $\mathcal{A}^{(r)} : V \times V \rightarrow \mathds{R}$:
\begin{equation}
  \label{eq:s-plus}
  \begin{split}
     & \mscore(G) = \\
     & = \exp{}\left(\boldsymbol{\theta}^T \mathbf{f}(G) + \sum_{r\in\mathcal{R}}\sum_{(s,t)\in E_r} \mathcal{A}^{(r)}(s,t)\right).
  \end{split}
\end{equation}

The association operator generalizes various models from the relation prediction literature:

\begin{description}
\item[TransE] \cite{bordes2013translating} embeds each relation $r$ into a vector in the shared space, representing a `difference' between sources and targets, to compute the association score under a translational objective,
$$\mathcal{A}_{\textsc{TransE}}^{(r)}(s,t) = - \Vert \ve_s + \ve_r - \ve_t \Vert . $$
\item[BiLin] \cite{nickel2011three} embeds relations into full-rank matrices, computing the score by a bilinear multiplication,
$$ \mathcal{A}_{\textsc{BiLin}}^{(r)}(s,t) = \ve_s^T \mathbf{W}_r \ve_t . $$
\item[DistMult] \cite{yang2014embedding} is a special case of \textbf{BiLin} where the relation matrices are diagonal, reducing the computation to a ternary dot product,
\end{description}
$$ ~~~\mathcal{A}_{\textsc{DistMult}}^{(r)}(s,t) = \langle \ve_s, \ve_r, \ve_t \rangle = \sum_{i=1}^{d} e_{s_i}~e_{r_i}~e_{t_i}. $$

\subsection{Parameter Estimation}
\label{ssec:learn}

The probabilistic formulation of \ergm{} requires the computation of a normalization term that sums over all possible graphs with a given number of nodes, $\mathcal{G}_N$. The set of such graphs grows at a rate that is super-exponential in the number of nodes, making exact computation intractable even for networks that are orders of magnitude smaller than semantic graphs like \WN. One solution is to approximate probability using a variant of the Monte Carlo Maximum Likelihood Estimation (MCMLE) produce,
\begin{equation}
  \label{eq:mcmle}
  \log P(G) \approx \log \psi(G) - \log \frac{|\mathcal{G}_{|V|}|}{M} \sum_{\Gneg \sim \mathcal{G}_{|V|}}^{M} \psi(\Gneg),
\end{equation}
where $M$ is the number of networks $\Gneg$ sampled from $\mathcal{G}_{|V|}$, the space of all (multirelational) edge sets on nodes $V$. Each $\Gneg$ is referred to as a \emph{negative sample}, and the goal of estimation is to assign low scores to these samples, in comparison with the score assigned to the observed network $G$.

Network samples can be obtained using edge-wise negative sampling.
For each edge $s\xrightarrow{r}t$ in the training network $G$, we remove it temporarily and consider $T$ alternative edges, keeping the source $s$ and relation $r$ constant, and sampling a target $\tneg$ from a \textbf{proposal distribution} $Q$. Every such substitution produces a new graph $\Gneg$,
\begin{align}
  \Gneg = &{} G \cup \{s\xrightarrow{r}\tneg\} \setminus  \{s\xrightarrow{r}t\}.
\end{align}

\paragraph{Large-margin objective.}
Rather than approximating the log probability, as in MCMLE estimation, we propose a margin loss objective: the log score for each negative sample $\Gneg$ should be below the log score for $G$ by a margin of at least 1. This motivates the hinge loss,
\begin{align}
  \notag
  \mathcal{L}(\boldsymbol{\Theta}, \Gneg; G) = \Big( 1 & - \log \mscore(G)\\
  &+ \log \mscore(\Gneg)\Big)_{+},
    \label{eq:loss}
\end{align}
where $(x)_+ = \max(0, x)$. Recall that the scoring function $\mscore$ includes both the local association score for the alternative edge and the global graph features for the resulting graph. However, it is not necessary to recompute all association scores; we need only subtract the association score for the deleted edge $s \xrightarrow{r} t$, and add the association score for the sampled edge $s \xrightarrow{r}\tneg.$

The overall loss function is the sum over $N=|E|\times T$ negative samples, $\{ \Gneg^{(i)}\}_{i=1}^N$, plus an $L_2$ regularizer on the model parameters,
\begin{equation}
  \label{eq:tot-loss}
  \mathcal{L}(\boldsymbol{\Theta}; G) =
  \lambda || \boldsymbol{\Theta} ||_2^2 + \sum_{i=1}^N \mathcal{L}(\boldsymbol{\Theta}, \Gneg^{(i)}).
\end{equation}

\paragraph{Proposal distribution.} The proposal distribution $Q$ used to sample negative edges is defined to be proportional to the local association scores
of edges not present in the training graph:
\begin{equation}
  \begin{split}
	Q(\tneg \mid s,r,G) \propto
    \begin{cases}
		0 & s\xrightarrow{r}\tneg~\in G \\
        \mathcal{A}^{(r)}(s,\tneg) & s\xrightarrow{r}\tneg~\notin G~.
	\end{cases} \\
  \end{split}
\end{equation}
By preferring edges that have high association scores, the negative sampler helps push the \sysname{} parameters away from likely false positives.

\section{Relation Prediction}
\label{sec:exp}

We evaluate \sysname{} on the relation graph edge prediction task.\footnote{Sometimes referred to as Knowledge Base Completion, e.g. in \newcite{SocherChenManningNg2013}.}
Data for this task consists of a set of labeled edges, i.e. tuples of the form $( s, r, t )$, where $s$ and $t$ denote source and target entities,
respectively.
Given an edge from an evaluation set, two prediction instances are created by hiding the source and target side, in turn.
The predictor is then evaluated on its ability to predict the hidden entity, given the other entity and the relation type.\footnote{We follow prior work in excluding the following from the ranked lists: the known entity (no self loops); entities from the training set which fit the instance; other entities in the evaluation set.}

\subsection{WN18RR Dataset}
\label{ssec:wnrr}

A popular relation prediction dataset for \WN{} is the subset curated as WN18 \cite{bordes2013translating,bordes2014semantic}, containing 18 relations for about 41,000 synsets extracted from \WN{} 3.0.
It has been noted that this dataset suffers from considerable leakage: edges from reciprocal relations such as hypernym / hyponym appear in one direction in the training set and in the opposite direction in dev / test  \cite{SocherChenManningNg2013,dettmers2018conve}.
This allows trivial rule-based baselines to achieve high performance.
To alleviate this concern, \newcite{dettmers2018conve} released the \textbf{WN18RR} set, removing seven relations altogether.
However, even this dataset retains four symmetric relation types: \textit{also see}, \textit{derivationally related form}, \textit{similar to}, and \textit{verb group}.
These symmetric relations can be exploited by defaulting to a simple rule-based predictor.

\subsection{Metrics}
We report the following metrics, common in ranking tasks and in relation prediction in particular:
\textbf{MR}, the Mean Rank of the desired entity;
\textbf{MRR}, Mean Reciprocal Rank, the main evaluation metric;
and \textbf{H@$k$}, the proportion of Hits (true entities) found in the top $k$ of the lists, for $k\in\{1,10\}$.
Unlike some prior work, we do not type-restrict the possible relation predictions (so, e.g., a \textit{verb group} link may select a noun, and that would count against the model).

\subsection{Systems}

We evaluate a single-rule baseline, three association models, and two variants of the \sysname{} re-ranker trained on top of the best-performing association baseline.

\subsubsection{\textsc{Rule}} We include a single-rule baseline that predicts a relation between $s$ and $t$ in the evaluation set if the same relation was encountered between $t$ and $s$ in the training set.
All other models revert to this baseline for the four symmetric relations.

\subsubsection{Association Models}
The next group of systems compute local scores for entity-relation triplets.
They all encode entities into embeddings $\ve$.
Each of these systems, in addition to being evaluated as a baseline, is also used for computing association scores in \sysname{}, both in the proposal distribution (see \Cref{ssec:learn}) and for creating lists to be re-ranked (see below): \textbf{\textsc{TransE}}, \textbf{\textsc{BiLin}}, \textbf{\textsc{DistMult}.}
For detailed descriptions, see \Cref{ssec:local}.

\subsubsection{Max-Margin Markov Graph Model}
The \sysname{} is applied as a re-ranker.
For each relation and source (target), the top $K$ candidate targets (sources) are retrieved based on the local association scores.
Each candidate edge is introduced into the graph, and the score $\mscore(G)$ is used to re-rank the top-$K$ list.

We add a variant to this protocol where the graph score and association score are weighted by $\alpha$ and $1-\alpha$, repsectively, before being summed.
We tune a separate $\alpha_r$ for each relation type, using the development set's mean reciprocal rank (MRR).
These hyperparameter values offer further insight into where the \sysname{} signal benefits relation prediction most (see \Cref{sec:analysis}).

Since we do not apply the model to the symmetric relations (scored by the \textsc{Rule} baseline), they are excluded from the sampling protocol described in \cref{eq:loss}, although their edges do contribute to the combinatory graph feature vector $\mathbf{f}$.

Our default setting backpropagates loss into only the graph weight vector $\boldsymbol\theta$.
We experiment with a model variant which backpropagates into the association model and synset embeddings as well.

\subsection{Synset Embeddings}
\label{ssec:embs}

For the association component of our model, we require embedding representations for \WN{} synsets.
While unsupervised word embedding techniques go a long way in representing wordforms~\cite{collobert2011natural,mikolov2013efficient,pennington2014glove}, they are not immediately applicable to the semantically-precise domain of synsets.
We explore two methods of transforming pre-trained word embeddings into synset embeddings.

\paragraph{Averaging.}
A straightforward way of using word embeddings to create synset embeddings is to collect the words representing the synset as surface form within the \WN{} dataset and average their embeddings~\cite{SocherChenManningNg2013}.
We apply this method to pre-trained GloVe embeddings~\cite{pennington2014glove} and pre-trained FastText embeddings~\cite{bojanowski2017enriching}, averaging over the set of all wordforms in all lemmas for each synset, and performing a case-insensitive query on the embedding dictionary.
For example, the synset `\textit{determine.v.01}' lists the following lemmas: `determine', `find', `find\_out', `ascertain'.
Its vector is initialized as $$\frac{1}{5}(\ve_{determine} + 2\cdot \ve_{find} + \ve_{out} + \ve_{ascertain}).$$

\paragraph{AutoExtend retrofitting + Mimick.}
AutoExtend is a method developed specifically for embedding \WN{} synsets~\cite{Rothe2015AutoExtend}, in which pre-trained word embeddings are retrofitted to the tripartite relation graph connecting wordforms, lemmas, and synsets.
The resulting synset embeddings occupy the same space as the word embeddings.
However, some \WN{} senses are not represented in the underlying set of pre-trained word embeddings.\footnote{We use the out-of-the-box vectors supplied in~\url{http://www.cis.lmu.de/~sascha/AutoExtend}.}
To handle these cases, we trained a character-based model called \textsc{Mimick}, which learns to predict embeddings for out-of-vocabulary items based on their spellings~\cite{pinter2017mimicking}.
We do not modify the spelling conventions of \WN{} synsets before passing them to Mimick, so e.g. \textit{`mask.n.02'} (the second synset corresponding to `mask' as a noun) acts as the input character sequence as is.

\paragraph{Random initialization.} In preliminary experiments, we attempted training the association models using randomly-initialized embeddings.
These proved to be substantially weaker than distributionally-informed embeddings and we do not report their performance in the results section.
We view this finding as strong evidence to support the necessity of a distributional signal in a type-level semantic setup.

\subsection{Setup}

Following tuning experiments, we train the association models on synset embeddings with $d=300$, using a negative log-likelihood loss function over 10 negative samples and iterating over symmetric relations once every five epochs.
We optimize the loss using AdaGrad with $\eta=0.01$, and perform early stopping based on the development set mean reciprocal rank.
\sysname{} is trained in four epochs using AdaGrad with $\eta=0.1$.
We set \sysname's re-rank list size $K=100$ and, following tuning, the regularization parameter $\lambda=0.01$ and negative sample count per edge $T=10$. Our models are all implemented in DyNet~\cite{dynet}.

\section{Results}
\label{sec:res}

\begin{table}
  \centering
  \small
  \begin{tabular}{llcccc}
    \toprule
    & System & MR & MRR & H@10 & H@1 \\
    \midrule
     & \textsc{Rule} & 13396 & 35.26 & 35.27 & 35.23 \\[6pt]
    1 & \textsc{DistMult} & 1111 & 43.29 & 50.73 & 39.67 \\
%     \textsc{Diag-r1} & 1672 & 43.40 & 50.94 & 39.73 \\
    2 & \textsc{BiLin} & \textbf{738} & 45.36 & 52.93 & 41.37\\
    3 & \textsc{TransE} & 2231 & \textbf{46.07} & \textbf{55.65} & \textbf{41.41} \\[6pt]
%     4 & \textsc{TransE}\textsubscript{AE+M} & 3344 &  44.46 & 51.75 & 40.90 \\[6pt]
%     5 & \sysname\textsubscript{Fine-Tune} & 1518 & 46.31 & 54.96 & 41.94 \\ % transe model after this: 1519 & 45.03 & 54.15 & 40.71 \\
    4 & \sysname & 2231 & 47.94 & \textbf{57.72} & 43.26 \\
    5 & \sysname\textsubscript{$\alpha_r$} & 2231 & \textbf{48.30} & 57.59 & \textbf{43.78} \\
    \bottomrule
  \end{tabular}
  \caption{\label{tab:dev} Results on development set (all metrics except MR are x100).
  \sysname{} lines use \textsc{TransE} as their association model.
  In \sysname\textsubscript{$\alpha_r$}, the graph component is tuned post-hoc against the local component per relation.}
\end{table}

\Cref{tab:dev} presents the results on the development set.
Lines 1-3 depict the results for local models using averaged FastText embedding initialization, showing that the best performance in terms of MRR and top-rank hits is achieved by \textsc{TransE}.
Mean Rank does not align with the other metrics; this is an interpretable tradeoff, as both \textsc{BiLin} and \textsc{DistMult} have an inherent preference for correlated synset embeddings, giving a stronger fallback for cases where the relation embedding is completely off,
but allowing less freedom for separating strong cases from correlated false positives, compared to a translational objective.

\paragraph{Effect of global score.} There is a clear advantage to re-ranking the top local candidates using the score signal from the \sysname{} model (line 4).
These results are further improved when the graph score is weighted against the association component per relation (line 5). We obtain similar improvements when re-ranking the predictions from \textsc{DistMult} and \textsc{BiLin}.

The \sysname{} training procedure is not useful in fine-tuning the association model via backpropagation:
this degrades the association scores for true edges in the evaluation set, dragging the re-ranked results along with them to about a 2-point drop relative to the untuned variant.

\Cref{tab:test-res} shows that our main results transfer onto the test set, with even a slightly larger margin. This could be the result of the greater edge density of the combined training and dev graphs, which enhance the global coherence of the graph structure captured by \sysname{} features.
To support this theory, we tested the \sysname{} model trained on only the training set, and its test set performance was roughly one point worse on all metrics, as compared with the model trained on the training+dev data.

\begin{table}
  \centering
  \small
  \begin{tabular}{lcccc}
    \toprule
    System & MR & MRR & H@10 & H@1 \\
    \midrule
    %\textsc{Rule}$^\dagger$ & 13219 & 36 & 36 & 36 \\
    \textsc{Rule} & 13396 & 35.26 & 35.26 & 35.26 \\[6pt]
    \textsc{ComplEx}$^\dagger$ & 5261 & 44\phantom{.00} & 51\phantom{.00} & 41\phantom{.00} \\
    \textsc{ConvE}$^\dagger$ & 5277 & 46\phantom{.00} & 48\phantom{.00} & 39\phantom{.00} \\ % v.5, published AAAI version
%     \textsc{ConvE}$^\dagger$ & 4187 & 43\phantom{.00} & 52\phantom{.00} & 40\phantom{.00} \\ % v.6, Latest on arXiv
    \textsc{ConvKB}$^\dagger$ & 2554 & 24.8\phantom{0} & 52.5\phantom{0} & \\[6pt]
    \textsc{TransE} & 2195 & 46.59 & 55.55 & 42.26 \\[6pt]
    \sysname\textsubscript{$\alpha_r$} & \textbf{2193} & \textbf{49.83} & \textbf{59.02} & \textbf{45.37}\\
    \bottomrule
  \end{tabular}
  \caption{\label{tab:test-res} Main results on test set.
  $^\dagger$ These models were not re-implemented, and are reported as in \newcite{kbc-cnn} and in
  \newcite{dettmers2018conve}.
  }
\end{table}

\paragraph{Synset embedding initialization.}
We trained association models initialized on AutoExtend+Mimick vectors (see \Cref{ssec:embs}).
Their performance, inferior to averaged FastText vectors by about 1-2 MRR points on the dev set, is somewhat at odds with findings from previous experiments on \WN~\cite{guu-miller-liang:2015:EMNLP}.
We believe the decisive factor in our result is the size of the training corpus used to create FastText embeddings, along with the increase in resulting vocabulary coverage.
Out of 124,819 lemma tokens participating in 41,105 synsets, 118,051 had embeddings available (94.6\%; type-level coverage 88.1\%). Only 530 synsets (1.3\%) finished this initialization process with no embedding and were assigned random vectors.
AutoExtend, fit for embeddings from~\newcite{mikolov2013efficient} which were trained on a smaller corpus, offers a weaker signal:
13,377 synsets (32\%) had no vector and needed Mimick initialization.

\section{Graph Analysis}
\label{sec:analysis}

\begin{table}
  \centering
  \small
  \begin{tabular}{lp{.9\linewidth}}
    \toprule
    \multicolumn{2}{l}{Positive}\\[4pt]
    1 & $s \xrightarrow{member\_meronym}{\circled{t}}$ \\
    2 & $s \xrightarrow[]{has\_part}{\circled{t}}$ \\
    3 & $s \xrightarrow[]{hypernym}{\circled{t}} \xrightarrow[]{derivationally\_related\_form}{u}$ \\[8pt]
    \multicolumn{2}{l}{Negative}\\[4pt]
    4 & $s \xrightarrow[]{hypernym}{\circled{t}}$ \\
    5 & $\circled{s} \xleftrightarrow[]{hypernym}{\circled{t}}$ \\
    6 & $s \xrightarrow[]{member\_meronym}{\circled{t}} \xrightarrow[]{instance\_hypernym}{u}$ \\
    7 & $s_1 \xrightarrow[]{has\_part}{\circled{t}} \xleftarrow[]{verb\_group}{s_2}$ \\[12pt]
    \bottomrule
  \end{tabular}
  \caption{\label{tab:top-feats} Select heavyweight features (motifs) following best dev set training using \sysname{}.
  Circled nodes count towards the motif.}
\end{table}

As a consequence of the empirical experiment, we aim to find out what \sysname{} has learned about \WN.
\Cref{tab:top-feats} presents a sample of top-weighted motifs.
Lines 1 and 2 demonstrate that the model prefers a broad scattering of targets for the \textit{member\_meronym} and \textit{has\_part} relations\footnote{Example edges: `America' $\rightarrow$ `American', `face' $\rightarrow$ `mouth', respectively.}, which are flat and top-downwards hierarchical, respectively, while line 4 shows that a multitude of unique \textit{hypernym}s is undesired, as expected from a bottom-upwards hierarchical relation.
Line 5 enforces the asymmetry of the hypernym relation.

Lines 3, 6, and 7 hint at deeper interactions between the different relation types.
Line 3 shows that the model assigns positive weights to hypernyms which have derivationally-related forms, suggesting that the derivational equivalence classes in the graph tend to exist in the higher, more abstract levels of the hypernym hierarchy, as noted in \Cref{sec:feat-eng}.
Line 6 captures a semantic conflict: synsets located in the lower, specific levels of the graph can be specified either as \textit{instances} of abstract concepts\footnote{Example \textit{instance\_hypernym} edge: `Rome' $\rightarrow$ `national capital'.}, or as \textit{members} of less specific concrete classes, but not as both.
Line 7 may have captured a nodal property -- since \textit{part\_of} is a relation which holds between nouns, and \textit{verb\_group} holds between verbs, this negative weight assignment may be the manifestation of a part-of-speech uniqueness constraint.
In addition, in features 3 and 7 we see the importance of symmetric relations (here \textit{derivationally\_related\_form} and \textit{verb\_group}, respectively), which manage to be represented in the graph model despite not being directly trained on.

\begin{table}
  \small
  \centering
  \begin{tabular}{llll}
  \toprule
    Source & Relation & Correct & Outranking  \\
    & & target & local target(s) \\
    \midrule
    indian lettuce & \textit{hypernym} & herb & garden lettuce\\[3pt]
    austria & \textit{has\_part} & vienna & germany,\\
    & & & hungary, france, \\
    & & & european union\\
    \bottomrule
  \end{tabular}
  \caption{\label{tab:rerank} Successful \sysname{} re-ranking examples.}
\end{table}

\begin{table}
  \centering
  \small
  \begin{tabular}{lclc}
    \toprule
    Relation $r$ & $\alpha_r$ & Relation $r$ & $\alpha_r$ \\
    \midrule
    mem. of domain usage & 0.78 & hypernym & 0.64 \\
    mem. of domain region & 0.77 & domain topic of & 0.38 \\
    member meronym & 0.67 & has part & 0.33 \\
    instance hypernym & 0.65 \\
    \bottomrule
  \end{tabular}
  \caption{\label{tab:alphas} Graph score weights found for relations on the dev set. Zero means graph score is not considered at all for this relation, one means only it is considered.}
\end{table}

\Cref{tab:rerank} presents examples of relation targets successfully re-ranked thanks to these features.
The first false connection created a new unique hypernym, `garden lettuce', downgraded by the graph score through incrementing the count of negatively-weighted feature 4.
In the second case, `vienna' was brought from rank 10 to rank 1 since it incremented the count for the positively-weighted feature 2, whereas all targets ranked above it by the local model were already \textit{has\_part}-s, mostly of `europe'.

The $\alpha_r$ values weighing the importance of \sysname{} scores in the overall function, found per relation through grid search over the development set, are presented in \Cref{tab:alphas}.
It appears that for all but two relations, the best-performing model preferred the signal from the graph features to that from the association model ($\alpha_r > 0.5$).
Based on the surface properties of the different relation graphs, the decisive factor seems to be that \textit{synset\_domain\_topic\_of} and \textit{has\_part}  pertain mostly to very common concepts, offering good local signal from the synset embeddings, whereas the rest include many long-tail, low-frequency synsets that require help from global features to detect regularity.

\pagebreak

\section{Conclusion}
This paper presents a novel method for reasoning about semantic graphs like \WN, combining the distributional coherence between individual entity pairs with the structural coherence of network motifs.
Applied as a re-ranker, this method substantially improves performance on link prediction.
Our analysis of results from \Cref{tab:top-feats}, lines 6 and 7, suggests that adding graph motifs which qualify their adjacent nodes in terms of syntactic function or semantic category may prove useful.

From a broader perspective, \sysname{} can do more as a probabilistic model than predict individual edges.
For example, consider the problem of linking a new entity into a semantic graph, given only the vector embedding.
This task involves adding multiple edges simultaneously, while maintaining structural coherence.
Our model is capable of scoring bundles of new edges, and in future work, we plan to explore the possibility of combining \sysname{} with a search algorithm, to automatically extend existing knowledge graphs by linking in one or more new entities.

We also plan to explore multilingual applications.
To some extent, the structural parameters estimated by \sysname{} are not specific to English: for example, hypernymy cannot be symmetric in any language.
If the structural parameters estimated from English \WN{} are transferable to other languages, then the combination of \sysname{} and multilingual word embeddings could facilitate the creation and extension of large-scale semantic resources across many languages~\cite{fellbaum2012challenges,bond2013linking,lafourcade2007making}.

\section*{Acknowledgments}
We would like to thank the anonymous reviewers for their helpful comments. We discussed fast motif-counting algorithms with Polo Chau and Oded Green, and received early feedback from Jordan Boyd-Graber, Erica Briscoe, Martin Hyatt, Bryan Leslie Lee, Martha Palmer, and Oren Tsur.
This research was funded by the Defense Threat Research Agency under award HDTRA1-15-1-0019.

\bibliographystyle{acl_natbib.bst}
\bibliography{citations}

\end{document}